\documentclass[letterpaper, 10 pt, conference, nofonttune]{ieeeconf} 

\IEEEoverridecommandlockouts                              

\usepackage{amsthm}

\usepackage{subfigure}
\usepackage{xcolor}
\usepackage{times}
\usepackage{epsfig}
\usepackage{amsmath}
\usepackage{amssymb}
\usepackage{mathrsfs}
\usepackage[ruled,vlined,linesnumbered]{algorithm2e}
\usepackage{wrapfig}
\usepackage{multirow}
\usepackage{microtype}

\usepackage{hyperref}
\usepackage{lineno}

\newcommand{\GGG}{\mathcal{G}}

\newcommand{\EEE}{\mathcal{E}}

\newcommand{\VVV}{\mathcal{V}}

\newcommand{\NNN}{\mathcal{N}}

\newcommand{\hh}{\mathbf{h}}

\newcommand{\vv}{\mathbf{v}}

\newcommand{\aaa}{\mathbf{a}}

\newcommand{\WW}{\mathbf{W}}

\newcommand{\zz}{\mathbf{z}}

\newcommand{\qq}{\mathbf{q}}

\newcommand{\pos}{\mathbf{p}}
\newcommand{\goal}{\mathbf{g}}

\makeatother

\title{\LARGE \bf
Coordinated Multi-Robot Navigation with Formation Adaptation
}

\begin{document}
\author{Zihao Deng$^{1}$, Peng Gao$^{2}$, Williard Joshua Jose$^{1}$, Christopher Reardon$^{3}$,\\ Maggie Wigness$^{4}$, John Rogers$^{4}$, and Hao Zhang$^{1}$
\thanks{*This work was partially supported by  NSF CAREER Award IIS-2308492, DARPA Young Faculty Award (YFA) D21AP10114-00,
and DEVCOM ARL TBAM CRA W911NF2520024.}
\thanks{$^{1}$Zihao Deng, Williard Joshua Jose, and Hao Zhang are with the Human-Centered Robotics Lab, University of Massachusetts Amherst, Amherst, MA 01002, USA.
Email: \{zihaodeng, wjose, hao.zhang\}@umass.edu.}%
\thanks{$^{2}$Peng Gao is with North Carolina State University, Raleigh, NC, 27695, USA. {Email: pgao5@ncsu.edu}.}%
\thanks{$^{3}$Christopher Reardon is with the University of Denver, Denver, CO 80208. Email: christopher.reardon@du.edu.}
\thanks{$^{4}$Maggie Wigness and John Rogers are with the U.S. Army DEVCOM Army Research Laboratory (ARL), Adelphi, MD 20783, USA. Email: \{maggie.b. wigness, john.g.rogers59\}.civ@army.mil.}%
}

\maketitle
\thispagestyle{empty}
\pagestyle{empty}

\begin{abstract}

Coordinated multi-robot navigation is an essential ability for a team of robots operating in diverse environments. Robot teams often need to maintain specific formations, such as wedge formations, to enhance visibility, positioning, and efficiency during fast movement. However, complex environments such as narrow corridors challenge rigid team formations, which makes effective formation control difficult in real-world environments. 
To address this challenge, we introduce a novel Adaptive Formation with Oscillation Reduction (AFOR) approach to improve coordinated multi-robot navigation.
We develop AFOR under the theoretical framework of hierarchical learning and integrate a spring-damper model with hierarchical learning
to enable both team coordination and individual robot control. 
At the upper level, a graph neural network facilitates formation adaptation and information sharing among the robots. At the lower level, reinforcement learning enables each robot to navigate and avoid obstacles while maintaining the formations.
We conducted extensive experiments using Gazebo in the Robot Operating System (ROS), a high-fidelity Unity3D simulator with ROS, and real robot teams. Results demonstrate that AFOR enables smooth navigation with formation adaptation in complex scenarios and outperforms previous methods.



More details of this work are provided on the project website: 
\url{https://hcrlab.gitlab.io/project/afor}.

\end{abstract}
\color{black}

\section{Introduction}
Multi-robot systems have become a promising solution for addressing complex tasks across a wide range of real-world applications, including search and rescue \cite{drew2021multi,queralta2020collaborative},
intelligent transportation \cite{hu2022multi, gao2023collaborative}, 
and space exploration \cite{han2020cooperative,indelman2018cooperative}.
These scenarios demand that multi-robot teams demonstrate high levels of coordination and synchronization to operate efficiently and ensure safety. 
A key capability in achieving these objectives is coordinated navigation, which enables robots to traverse in a coordinated manner as a team in the environment. In such scenarios, robots often need to maintain specific formations for optimal performance. 
For example, wedge formations are commonly used by human teams and bird flocks to achieve a balance between visibility, offense, defense, and efficiency during high-speed movement.

\begin{figure}[!t]
\centering
\vspace{6pt}
\includegraphics[width=0.475\textwidth]{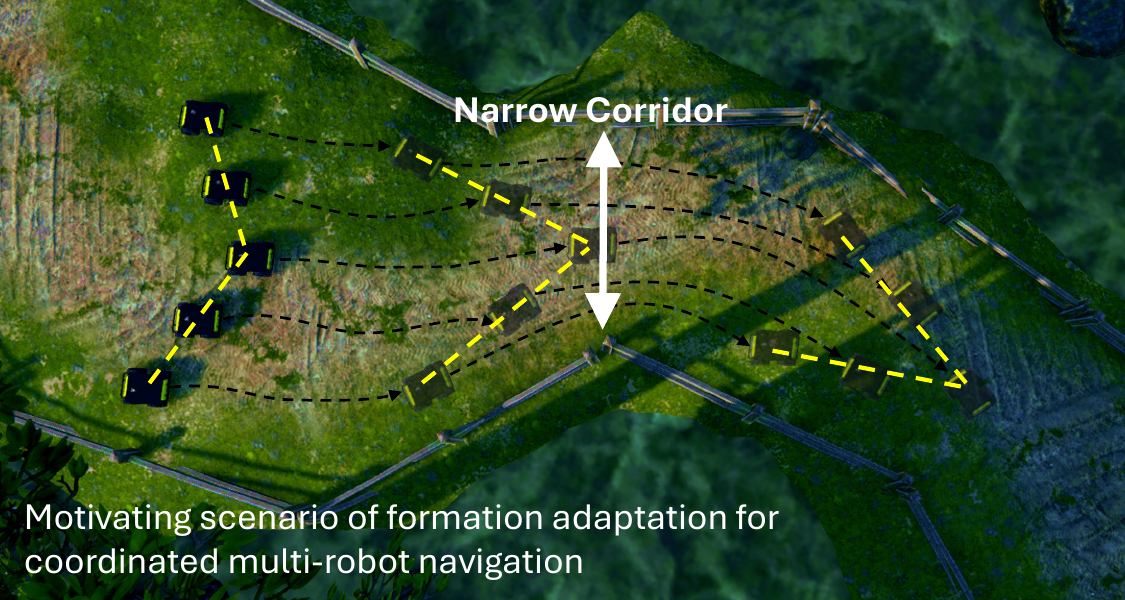}
\vspace{-6pt}
\caption{A motivating scenario for coordinated multi-robot navigation with formation adaptation. 
The team of robots must dynamically adapt their desired formation to navigate through a narrow bridge 
while smoothly moving towards their goal positions.}
\label{fig:tbam-motivation}
\vspace{-12pt}
\end{figure}

However, real-world environments are often complex and include narrow corridors, where rigid team formations cannot maneuver effectively. For example, as shown in Figure \ref{fig:tbam-motivation}, a team of robots in a rigid wedge formation cannot go through a narrow bridge that cannot accommodate the full width of the formation. 
In such situations, robots must dynamically adjust their formations to adapt to environmental constraints while navigating toward their goal positions. Additionally, oscillating robot motions  can affect both efficiency and safety, which must be addressed to enable smooth coordinated navigation.

Due to the importance of coordinated multi-robot navigation, various methods have been implemented. Learning-free approaches, such as game theory \cite{tang2020multi}, traditional path planning \cite{galceran2013survey}, and optimization techniques \cite{thabit2018multi,reily2020leading}, are commonly used,
but often struggle with adaptability in complex environments and tend to be computationally intensive.
Recently, learning-based methods have demonstrated improved performance. Reinforcement learning approaches for multi-robot coordination \cite{chen2017socially, yang2020multi} utilize control policies approximated by neural networks \cite{li2020graph, zhang2023neural}. However, the previous methods generally cannot address 
formation adaptation for a team of robots to adapt to narrow corridors during coordinated navigation.


To address these difficulties, we introduce a novel \textit{Adaptive Formation with Oscillation Reduction} (AFOR) approach for coordinated multi-robot navigation, 
which enables a team of robots to adapt their formations for navigating narrow corridors while maintaining a more stable team formation with reduced oscillations. 
Based upon a graph representation to encode the robot team, AFOR is developed in the theoretical framework of hierarchical learning together with a spring-damper model to enable both team decision making and individual robot control. 
At the upper level, a graph neural network is used to facilitate information sharing and coordination among robots.
At the lower level, AFOR uses reinforcement learning to learn individual robot control,
which enables each robot to execute the team coordination and avoid obstacles.




Our key contribution is the introduction of a novel AFOR approach to enable coordinated multi-robot navigation with formation adaptation. Two specific novelties include:
\begin{itemize}
    \item This work introduces one of the first learning solutions for formation adaptation with oscillation reduction, which enables the new team capability for robots to adapt their formations to navigate complex environments, particularly narrow corridors, while moving more stably toward their goal positions.
    
    \item We introduce a novel hierarchical learning method that incorporates a spring-damper model to simultaneously learn team decision-making for coordination at the upper level and individual robot control for navigation and obstacle avoidance at the lower level. A unified learning algorithm based on Proximal Policy Optimization (PPO) is also developed to train this hierarchical approach.
\end{itemize}
As a practical contribution, we conduct extensive evaluation and validation of AFOR using a standard Gazebo simulation in ROS, a high-fidelity Unity3D simulator with ROS, and real robot teams.

\color{black}
\section{Related Work}\label{sec:related}

\subsection{Learning-Free Coordinated Multi-Robot Navigation}
From a theoretical perspective, traditional methods of coordinated multi-robot navigation are categorized into three groups: classic, game theory, and optimization techniques. 
Classic graph-based methods normally include algorithms such as A$^*$ \cite{he2022dynamic, sun2019novel}, Dijkstra \cite{bai2019distributed, mac2017hierarchical}, and D$^*$ \cite{peng2015multi}. 
Game theory provides a framework for multi-robot to make optimized decisions by considering the potential decisions of other robots \cite{tang2020multi}, \cite{cappello2020hybrid}. 
Additionally, optimization-based methods focus on utilizing mathematical optimization for positioning robots. Graph-based optimization methods allow robots to find a safe global path by generating a graph of feasible formations of robots through sampling convex regions in free space \cite{alonso2017multi}, \cite{alonso2016distributed}. Hierarchical Quadratic Programming (HQP) represents another optimization method, involving multiple levels of priority among different tasks \cite{koung2021cooperative}. However, these methods are computationally intensive and often slow to respond to real-time environmental changes.

From a multi-robot formation control perspective, there are two primary approaches: leader-follower and virtual structure methods. In the leader-follower approach, a designated leader guides the movement of follower robots, simplifying control but creating a high dependency on the leader \cite{di2021multi, yu2019formation, reily2020leading, wu2022leader, xiao2019leader}. Some variations incorporate artificial potential fields \cite{ying2015leader} or learning frameworks \cite{bai2021learning}. In contrast, virtual structure methods maintain formation by keeping robots in fixed positions within a predefined structure \cite{benzerrouk2014stable, alonso2019distributed, roy2019virtual, abujabal2023comprehensive, roy2018multi}. However, the previous methods above did not well address adaptive formation control, which is crucial for a team of robots to navigate through complex environments where the space is too narrow for the entire team to pass with a rigid formation. 

\subsection{Hierarchical Learning for Multi-Robot Navigation}


Recently, learning-based methods have shown promising results on coordinated navigation. Reinforcement learning (RL) based methods were implemented to address the problems of slow responses to environmental changes \cite{han2020cooperative}, \cite{han2022reinforcement}, \cite{hacene2021behavior}. However, it is common for these single-level learning methods to have non-converging outcomes.


To address this challenge, hierarchical learning has significantly enhanced the ability of adaptive systems to address complex tasks by incorporating multi-level learning processes. 
The lower level typically focuses on individual robot control, using RL for tasks like obstacle avoidance \cite{bischoff2013hierarchical,jin2021hierarchical}. 
Deep RL improves this by using neural networks for more complex environments and decisions \cite{yang2018hierarchical,chang2023hierarchical}.
The upper level in such systems is responsible for planning and coordination. For example, it selects sub-goals through goal-based planning \cite{wohlke2021hierarchies}, divides exploration areas using dynamic Voronoi partitions \cite{hu2020voronoi}, and facilitates safe obstacle avoidance \cite{zhu2022hierarchical}. Additionally, it enables efficient searching by narrowing down exploration spaces through the use of prior knowledge \cite{pal2021learning}. 
Building upon these concepts, neural networks have been proposed at the upper level to enhance decision making \cite{liu2020robot}. 
Moreover, advanced models like convolutional \cite{li2020graph} and graph neural networks \cite{blumenkamp2022framework} are developed to facilitate communication between robots, which helps them extract environmental features.



While previous research has shown the promise of learning methods for multi-robot navigation and formation control, the problem of formation adaptation has not been addressed yet.

\color{black}

\begin{figure*}[h]
\includegraphics[width=0.75\textwidth]{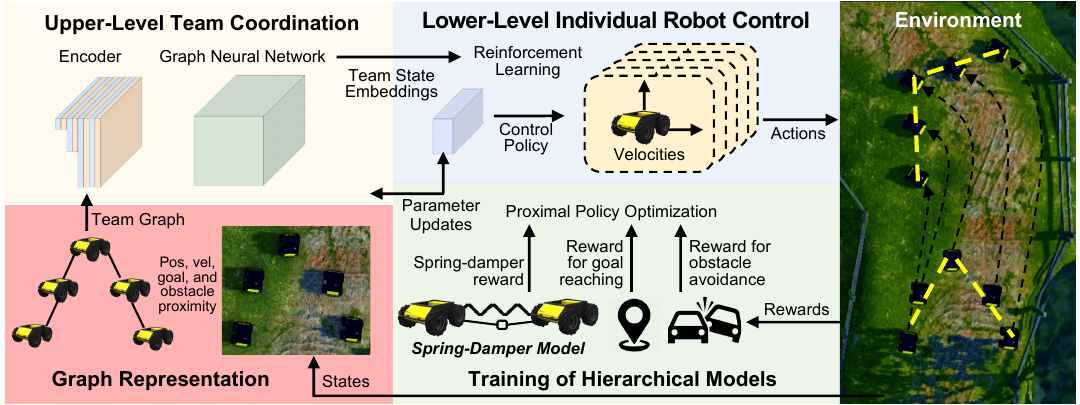}
\centering
\vspace{-6pt}
\caption{Overview of AFOR for coordinated multi-robot navigation with formation adaptation. 
AFOR's unified hierarchical learning model uses an upper-level GNN to share information and make team coordination decisions, while the lower-level RL learns individual robot control for navigation and obstacle avoidance.
AFOR also integrates a spring-damper model with the hierarchical learning to enable formation adaptation and reduce oscillations.}\label{fig:architecture}
\vspace{-9pt}
\end{figure*}


\section{Approach}\label{sec:approach}
\subsection{Problem Definition}

We represent a team of $n$ robots as a graph 
$\GGG=\{\VVV, \EEE\}$.
Each robot is represented as a node within the node set 
$\VVV=\{\vv_1,\vv_2,\dots,\vv_n\}$.
Each node $\vv_i = [\mathbf{p}_i, \mathbf{q}_i, \mathbf{g}_i]$ includes the attributes of the $i$-th robot associated with its position $\pos_i=[p_{x,i}, p_{y,i}]$, velocity $\qq_i=[q_{x,i}, q_{y,i}]$, and goal position $\goal_i=[g_{x,i}, g_{y,i}]$. 
We employ a radius graph to construct the edge set 
$\EEE=\{e_{i,j}\}$, where $e_{i,j} = 1$ if the $i$-th and $j$-th robot are within a radius, otherwise $e_{i,j} =0$.

Given the graph $\GGG$, we further define the state of each robot as a concatenation of its attributes $\mathbf{s}_i = [\mathbf{p}_i, \mathbf{q}_i, \mathbf{g}_i, d_i]$, where $d_i$ represents the distance from the nearest obstacle to the $i$-th robot. The action of each robot is defined as $\mathbf{a}_i = [v_{x,i}, v_{y,i}]$, where $v_{x,i}$ and $v_{y,i}$ denote the output velocity of the $i$-th robot in $x$ and $y$ directions, respectively. 
Our objective is to address the following adaptive formation control problem:
\begin{itemize}
    \item Formation Control: the coordination of multiple robots to move and maintain a specific geometric arrangement while navigating to the goal position together as a team.
    \item Formation Adaptation: 
    the new ability of a group of robots to dynamically adjust their relative positions in response to changes in the environment, while maintaining the desired formation to reach the goal position as a team. 
\end{itemize}

\color{black}

\subsection{Hierarchical Learning for Multi-Robot Formation Control}

In order to enable formation control for coordinated multi-robot navigation, 
we propose a novel AFOR with a hierarchical learning formulation, 
with the upper level to generate team decisions for robot coordination,
and the lower level to control individual robots for navigation and obstacle avoidance.
An overview of our AFOR approach is illustrated in Figure \ref{fig:architecture}.

\subsubsection{Coordination at the Upper Level}
AFOR's upper level aims to make team decisions to coordinate the robots and adapt formation controls according to the surrounding environment.
To achieve this goal,
we design a Graph Neural Network (GNN) as the upper-level learning of AFOR to encode complex spatial relationships among robots
and make team decisions facilitated by neural message passing. 
Formally, the state is computed by  $\zz_i = M\!L\!P(\mathbf{s}_i)$,
where $\zz_i$ denotes the embedding of the $i$-th robot state and  $M\!L\!P$ denotes a Multi-Layer Perceptron with four linear layers followed by ReLU activation. Given $\zz_i$, we further compute robot state embedding as follows:
\begin{equation}
    \hh_i = \WW^h \zz_i + \sum_{j \in \NNN(i)} \WW^h \left( \zz_j - \zz_i \right)
\end{equation}
where $\NNN(i)$ represents the set of neighboring robots to encode the messaging passing in GNN, and $\WW^h$ is the learnable weight matrix of the GNN. Intuitively, the GNN is designed to learn the state difference between the $i$-th robot and its neighboring robots, which is encoded in the aggregation $\zz_j - \zz_i$. 
The upper level does not only consider the state of each robot but also its neighboring teammates to encode information of the team, 
thus facilitating team coordination.

\subsubsection{Individual Robot Control at the Lower Level}
Given the state embedding $\hh_i$, we design a navigation control network that generates actions for each robot based on its embedding, $\mathbf{a}_i = \phi(\hh_i)$,
where $\phi$ represents the low-level network composed of two linear layers followed by ReLU activations. This network translates the state embeddings into specific actions $\aaa_i$ for each robot, which allows them to navigate efficiently toward their goals.  We employ reinforcement learning to govern the low-level control through a policy $\pi_{\theta}(\aaa_i|\mathbf{s}_i)$, parameterized by $\theta$, for  navigation and obstacle avoidance. 
This approach allows the robots to learn and optimize their navigation strategies by continuously interacting with the environment. 
As $\aaa_i$ are computed and executed, the robots engage both individually and as a coordinated team. The environment provides feedback in the form of rewards and updated states, which are captured and fed back into the system, which refines both high-level coordination and low-level control policies.


\subsubsection{Spring-Damper Model for Formation Adaptation}



While AFOR's hierarchical learning model enables multi-robot formation control,
robots may encounter scenarios where they cannot navigate effectively if they rigidly maintain their formation.
In addition, oscillation in the robots' movements may occur during navigation, due to frequent formation adjustments and inconsistencies in coordination between robots.

To address these challenges, we propose to integrate a spring-damper model with the hierarchical learning into AFOR. 
The spring-damper model simulates flexible connections between the robots in the team.
\begin{itemize}
    \item The spring component, inspired by the Hooke’s law \cite{rychlewski1984hooke}, flexibly maintains the desired distances between robots for formation control  
and ensure that the robots remain close enough to navigate tight spaces but far enough to avoid collisions. 
Mathematically, given a pair of robots $\mathbf{v}_i$ and $\mathbf{v}_j$, 
their desired distance to maintain a formation and their actual distance are denoted by $\mathbf{d}_{i,j}$ and $\mathbf{p}_{i,j} = ||\mathbf{p}_i-\mathbf{p}_j||_2$. 
Then, the spring component can be computed based on $|\mathbf{d}_{i,j} - \mathbf{p}_{i,j}|$.
    \item The damper component applies a force proportional to the robots' relative velocities, 
    which can be computed by $\mathbf{q}_{i,j} = |\mathbf{q}_{i}-\mathbf{q}_{j}|$.
    This damper component is designed to prevent overshoot and improves smooth navigation. 
\end{itemize}
In our hierarchical learning method, we model both components using a reward function defined as follows:
\begin{equation}
   R^\textit{adapt} = \sum_{\mathbf{v}_i, \mathbf{v}_j \in \mathcal{V}} -\alpha |\mathbf{d}_{i,j} - \mathbf{p}_{i,j}| - (1-\alpha)\mathbf{q}_{i,j}
\end{equation}
where $\alpha$ is a hyperparameter to balance the spring and damper components. 
This reward encoding the spring-damper model is used together with the classic rewards defined for 
reaching the goal position and obstacle avoidance for training AFOR.
A detailed description of the classic rewards is included in the supplemental document on the project website.

\begin{figure*}[htbp]
\centering
\subfigure[Circle Formation]{\includegraphics[height=3.95cm]{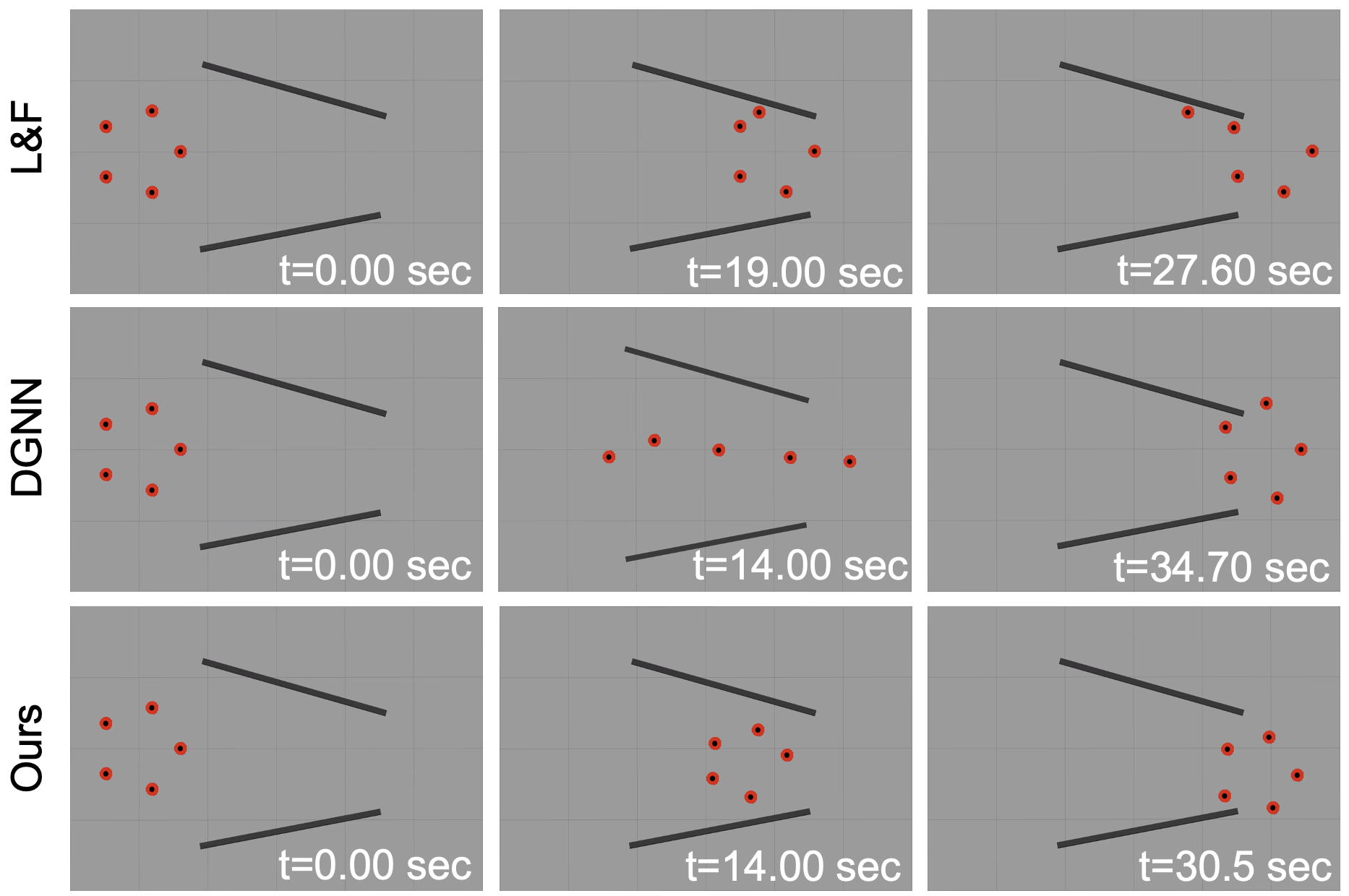}}\label{fig:qual_circle}
\subfigure[Wedge Formation]{\includegraphics[height=3.95cm]{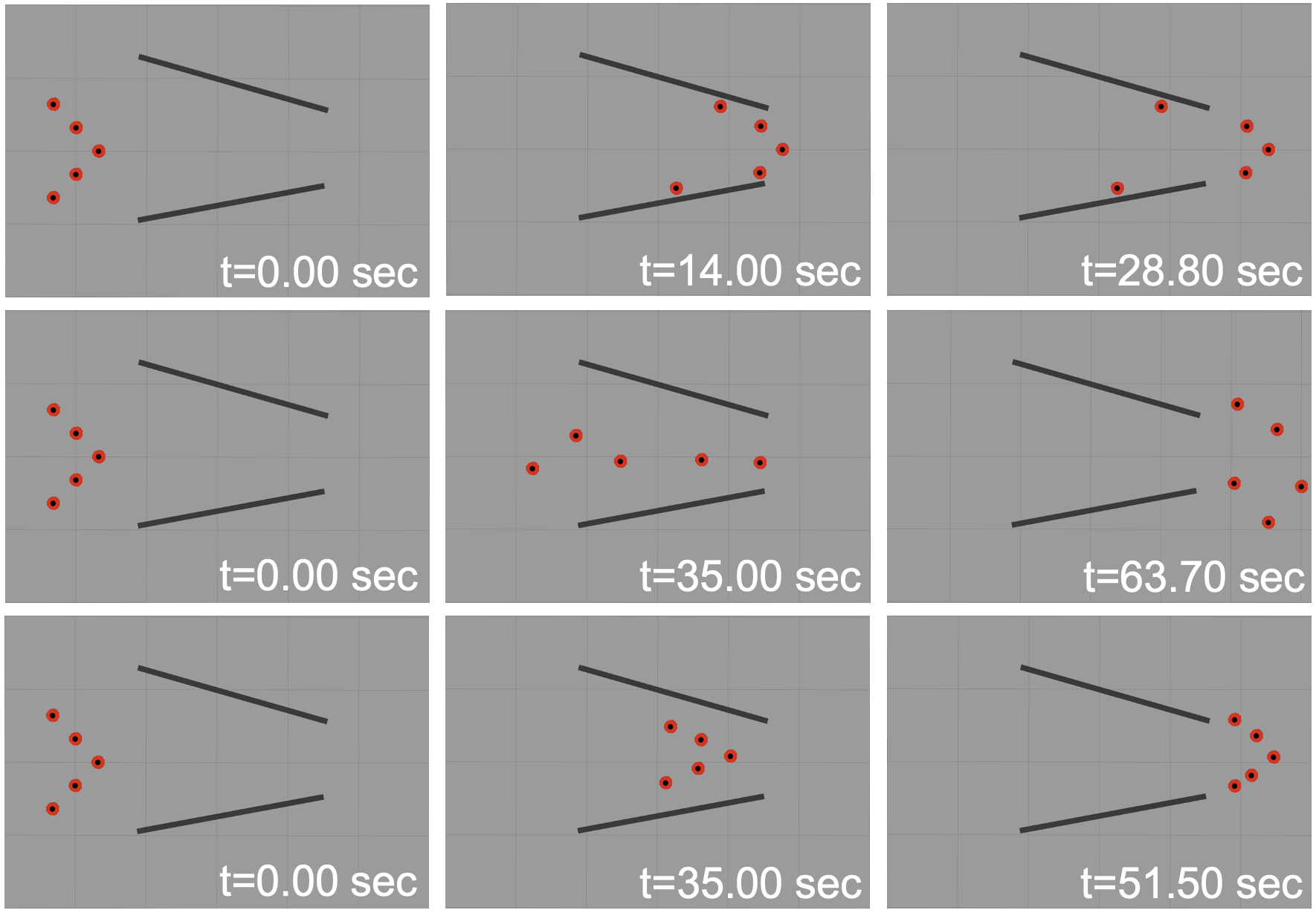}}\label{fig:qual_wedge}
\subfigure[Line Formation]{\includegraphics[height=3.95cm]{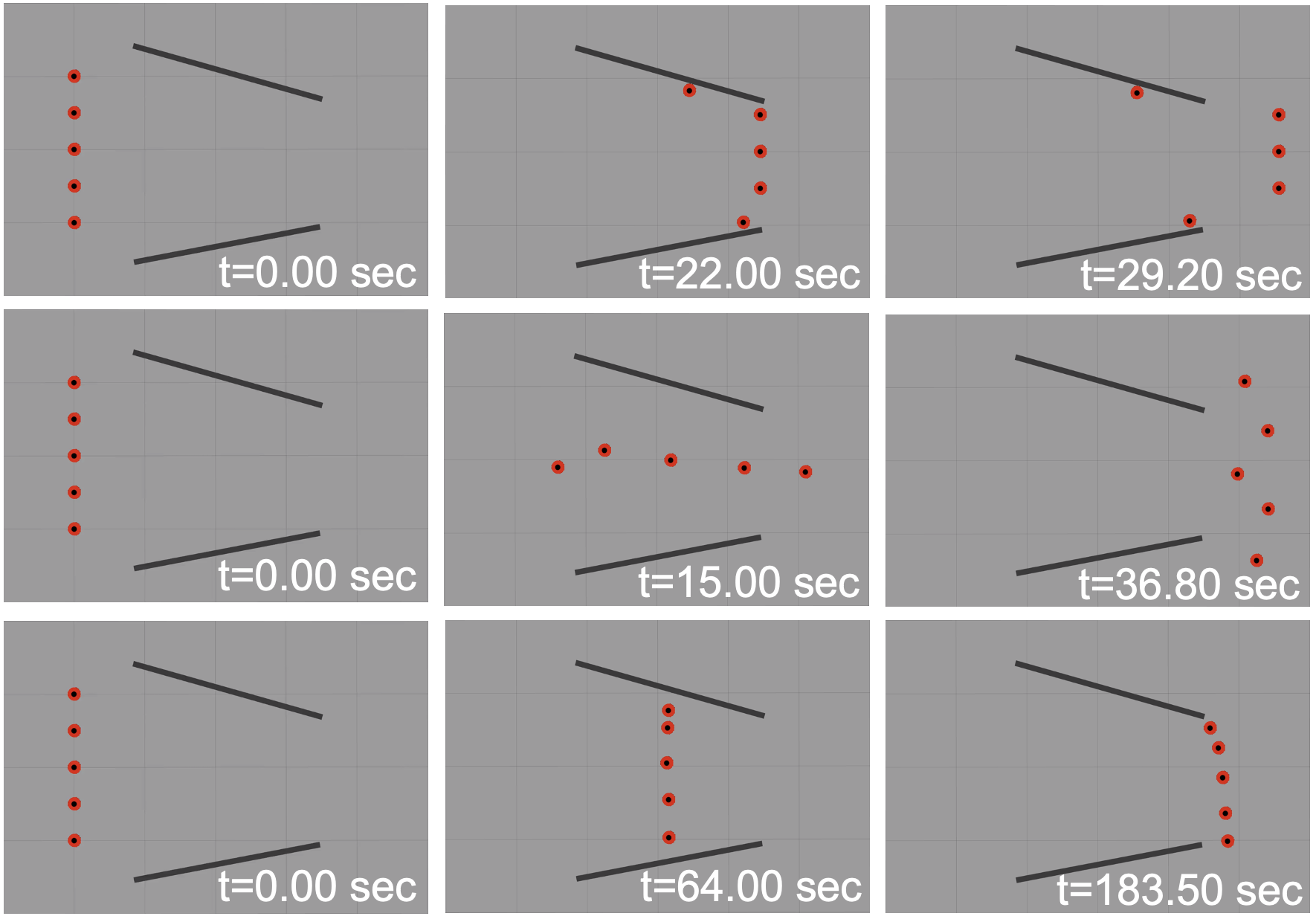}}\label{fig:qual_line}
\vspace{-6pt}
\caption{Qualitative results on coordinated navigation with robots maintaining circle, wedge and line formations in Gazebo simulations using ROS. 
}
\label{fig:qual}\vspace{-9pt}
\end{figure*}

\subsubsection{Training AFOR}

In order to train AFOR as a hierarchical learning model, we implement a PPO algorithm to synchronize the training of the upper for team decision making and lower level for individual robot control. 
The algorithm receives input during each training step from actions $\mathbf{a}_i$, states $\mathbf{s}_i$, and rewards of all robots. 
These rewards are then employed to compute the advantage estimates $A^{\pi_\textit{old}}(\mathbf{s}_i,\aaa_i)$, which represent how much better or worse taking $\aaa_i$ was, compared to the expected rewards under the old policy $\pi_{\theta_\textit{old}}(\mathbf{s}_i,\aaa_i)$. 
To compute a single loss for all robots, the updated policy $\pi_{\theta}(\mathbf{s}_i,\aaa_i)$, as the output of PPO, is compared with the old policy $\pi_{\theta_\textit{old}}(\mathbf{s}_i,\aaa_i)$. 
To ensure stable updates, a clipping function $\text{clip}(1-\lambda, 1+\lambda)$ is applied to limit the ratio of the updated and old policy. 
This ensures that policy updates are incremental and stay within a trust region defined by $\lambda$. 
Formally, the PPO loss is defined as follows:
\begin{equation}
\begin{split}
\sum_{\mathbf{v}_i \in \mathcal{V}} & \ \mathbb{E}_{\mathbf{s}_i, \aaa_i \sim d^{\pi_{\theta_\textit{old}}}}\left[\min\left(\frac{\pi_{\theta}(\aaa_i|\mathbf{s}_i)}{\pi_{\theta_\textit{old}}(\aaa_i|\mathbf{s}_i)} A^{\pi_\textit{old}}(\mathbf{s}_i,\aaa_i), \right. \right. \\
& \ \left.\left. \text{clip}\left(\frac{\pi_{\theta}(\aaa_i|\mathbf{s}_i)}{\pi_{\theta_\textit{old}}(\aaa_i|\mathbf{s}_i)}, 1-\lambda, 1+\lambda\right) A^{\pi_\textit{old}}(\mathbf{s}_i,\aaa_i)\right)\right]
\end{split}
\end{equation}
where $\mathbb{E}$ represents the expectation over the sampling distribution of $\aaa_i$ and states $\mathbf{s}_i$ from all robots during training, under the visitation distribution $d^{\pi_{\theta_\textit{old}}}$. 
This distribution defines the probability of encountering state $\mathbf{s}_i$ and taking action $\aaa_i$ while following the old policy $\theta_\textit{old}$. 

PPO optimizes the loss from sample trajectories collected by interacting with the environment.
The gradients computed from this loss are backpropagated through the entire hierarchical learning model, which updates both the upper and lower level parameters simultaneously. 
At the lower level, the gradients update the policy $\pi_{\theta}$ for individual robot control. 
At the upper level, the gradients update the weights $\WW^h$ of the GNN to generate team-level embeddings $\hh_i$ to facilitate team coordination.

\color{red}

\color{black}

\section{Experiments}\label{sec:experiment}
\subsection{Experimental Setups}



We evaluate our multi-robot coordinated navigation method with centralized execution, using a standard Gazebo simulation in ROS, a high-fidelity Unity3D simulator with ROS, and real robot teams. In each scenario, the team of robots needs to traverse a narrow corridor while maintaining their specific formations such as circle, wedge, or line. 
Details on approach implementation and training are included in the supplementary document on the project website.



\color{black}

\begin{table*}[ht]
\centering
\tabcolsep=0.065cm
\caption{Quantitative results with three formations in Gazebo execution.}
\label{tab:formation-integrity}
\vspace{-6pt}
\begin{tabular}{|l|c|c|c|c|c|c|c|c|c|c|c|c|c|c|c|}
\hline
Method  & \multicolumn{5}{c|}{Circle Formation} & \multicolumn{5}{c|}{Wedge Formation} &  \multicolumn{5}{c|}{Line Formation}  \\
\cline{2-16}
         & SR (\%) & $\delta<0.5$ & $\delta<0.3$ & $\delta<0.1$ & $\delta<0.03$
         & SR (\%) & $\delta<0.5$ & $\delta<0.3$ & $\delta<0.1$ & $\delta<0.03$
         & SR (\%) & $\delta<0.5$ & $\delta<0.3$ & $\delta<0.1$ & $\delta<0.03$  \\

\hline\hline
L$\&$F \cite{reily2020leading} & 80 & 75.114 & 69.107 & 70.105 & 66.043  
                               & 60 & 81.436 & 68.335 & 66.691 & 62.864  
                               & 60 & 63.758 & 59.861 & 55.588 & 55.588  \\
\hline
DGNN \cite{blumenkamp2022framework} & \textbf{100} & 60.413 & 59.330 & 58.906 & 58.906  
                                    & 60 & 47.850 & 44.935 & 42.325 & 41.917  
                                    & \textbf{100} & 27.895 & 22.122 & 20.155 & 20.155  \\
\hline
Baseline & \textbf{100} & 84.450 & 84.097 & 80.521 & 79.880  
   & \textbf{100} & 86.506 & 76.491 & 68.219 & 66.733  
   & \textbf{100} & 82.918 & 81.731 & 64.264 & 64.438  \\
\hline
AFOR (ours) & \textbf{100} & \textbf{92.887} & \textbf{90.891} & \textbf{90.397} & \textbf{88.662}  
     & \textbf{100} & \textbf{91.391} & \textbf{91.841} & \textbf{90.354} & \textbf{87.959}  
     & \textbf{100} & \textbf{88.596} & \textbf{84.525} & \textbf{85.131} & \textbf{72.901}  \\
\hline
\end{tabular}
\end{table*}

We compare our AFOR with three baseline and previous methods:  
(1) A Baseline method ($\textbf{BL}$) that uses hierarchical learning but excludes the integration of the spring-damper model.
(2) A Leader and Follower method ($\textbf{L\&F}$)  \cite{reily2020leading} that designates one robot within the team as the ``leader'' for guiding the movements and actions of the other robots, referred to as ``followers". 
(3) Decentralized GNN ($\textbf{DGNN}$) \cite{blumenkamp2022framework} that uses hierarchical learning to directly control individual robots for navigation, which does not incorporate team-level formation control or adaptation.

\begin{figure}[htbp]
\centering
\vspace{-6pt}
\subfigure[Circle Trajectory]{\includegraphics[height=3.6cm]
{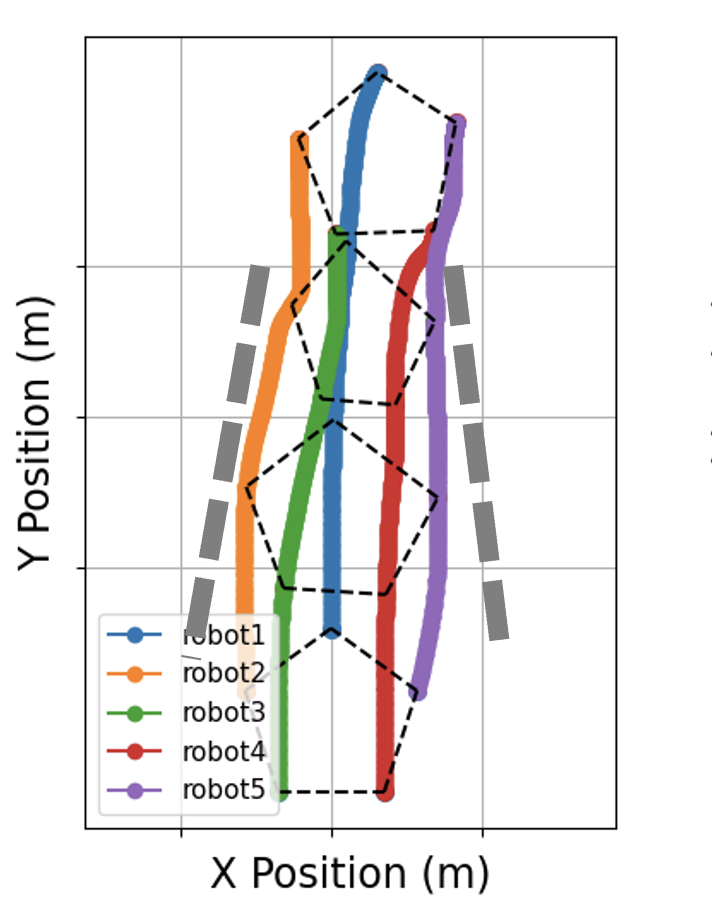}}\label{fig:cir_5}
\subfigure[Wedge Trajectory]{\includegraphics[height=3.6cm]
{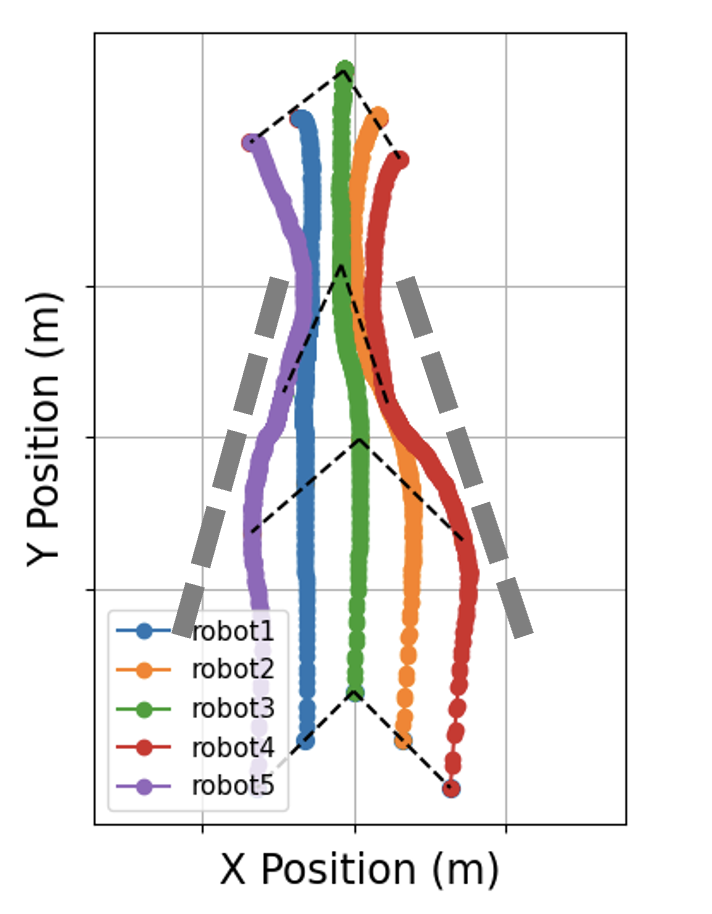}}\label{fig:wedge_5}
\subfigure[Line Trajectory]{\includegraphics[height=3.6cm]
{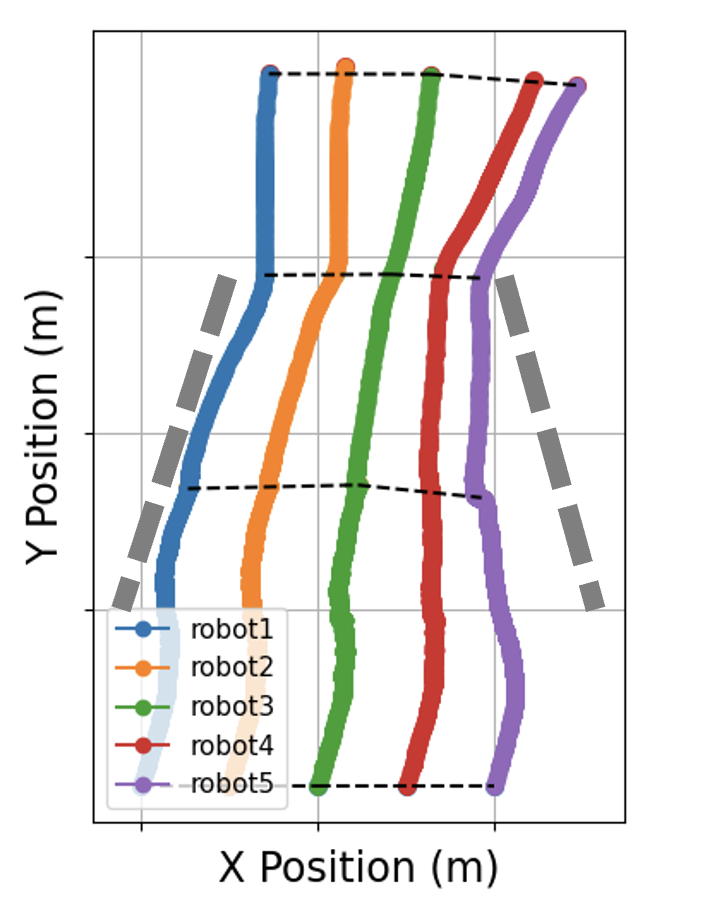}}\label{fig:line_5}
\vspace{-12pt}
\caption{
The trajectories depict a team of robots navigating a narrow corridor while adaptively maintaining formations. Each robot's path is represented by a different color, with key timestamps marking formation transitions. The gray lines indicate obstacles. 
}
\label{fig:trajectories}
\end{figure}

\begin{figure*}[htb]
\centering
\subfigure{\includegraphics[width=0.950\textwidth]{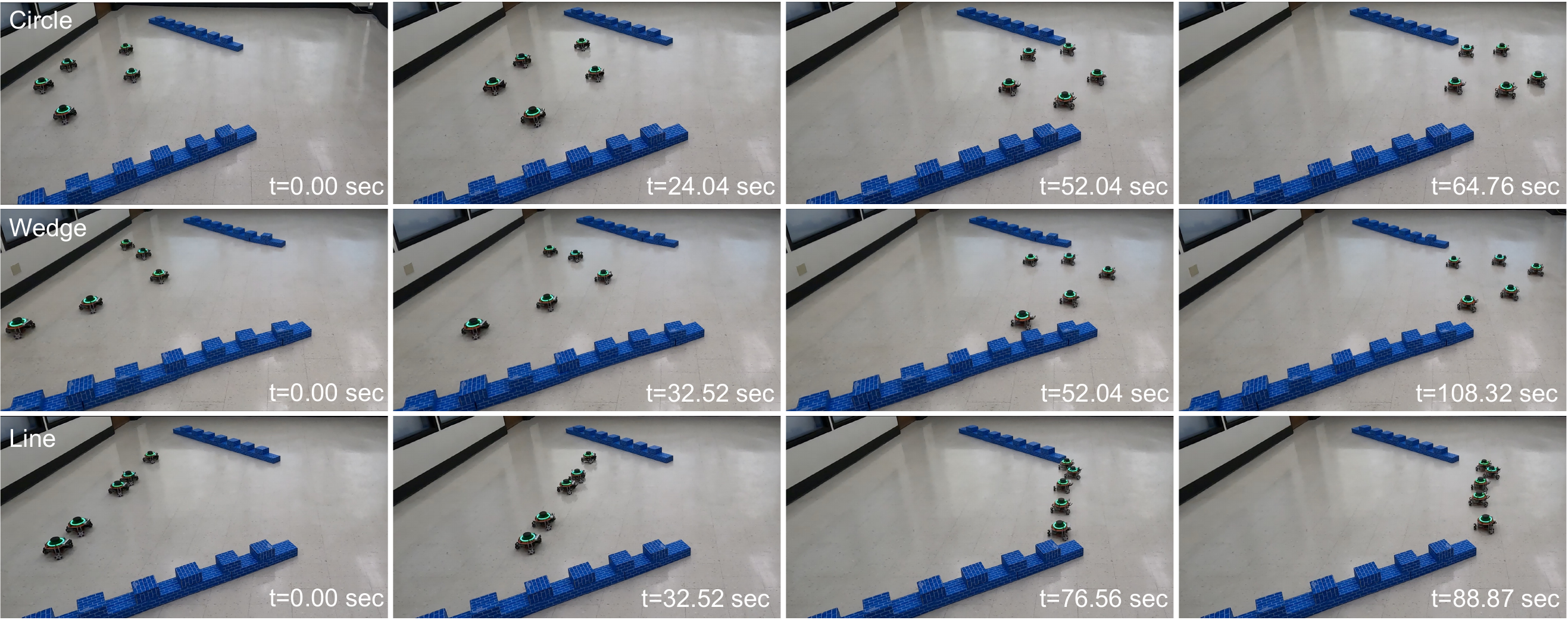}}
\vspace{-6pt}
\caption{Qualitative results on formation adaptation for coordinated multi-robot navigation using real mobile robots that form circle, wedge and line formations.}
\label{fig:real}
\end{figure*}

We employ two metrics for method evaluation and comparison:
(1) Successful Rate ($\textbf{SR}$) for navigation is defined as the proportion of the number of robots successfully reaching their destination without any collisions. 
(2) Contextual Formation Integrity ($\textbf{CFI}$) is defined by the real-time adherence of the robots to their optimal formation shape, measured as a percentage. The metric incorporates threshold and uncertainty concepts commonly used in computer vision \cite{Godard_2019_ICCV}.
Mathematically, CFI is defined and computed as follows:
\begin{equation}
\footnotesize
\beta\left(1 - \delta^{-1}\min\left( \left| W - (\tau + \delta) \right|, \left| W - (\tau - \delta) \right| \right)\right) 
+ \, (1-\beta) \epsilon
\nonumber
\end{equation}
The first term of CFI assesses the gap usage, where $W$ denotes the team formation's maximum width, $\tau$ is a safety threshold computed as the corridor width minus a safety margin,
and $\delta$ represents the rigidness of the formation, with a smaller value indicating stricter formation requirements. 
The second term $\epsilon \in [0,1]$ assesses the team's shape integrity.
Putting the two terms together, CFI evaluates how well a robot team utilizes the corridor space and maintains formation integrity, balanced by the coefficient $\beta \in [0,1]$.
 $\text{CFI}\in[0,1]$, with higher values indicating better performance. 
 In our experiments, we set $\tau$ to twice the robot's width and $\beta=0.5$.
 Examples of calculating CFI for the wedge, circle, and line formations are included in the supplemental document on the project website.




\subsection{Results on Formation Adaptation in Gazebo Simulations}
We first evaluate our approach in Gazebo simulations and explore the navigation capabilities of three distinct formations through a progressively narrowing corridor.


The qualitative results are shown in Figure \ref{fig:qual}. 
We observe that L$\&$F struggles in the narrow corridor in three different scenarios, becoming stuck due to its lack of adaptive formation control. 
DGNN is able to control the robot team to reach the destination; however, since it lacks multi-robot formation control capabilities, the team formation cannot be maintained in all scenarios.
By addressing both formation control and formation adaptation with oscillation reduction, our method not only successfully reaches the goal with fewer oscillations but also dynamically adjusts formations in response to environmental changes. This adaptive behavior enhances both space utilization and formation integrity, particularly in challenging scenarios like narrow passages, where rigid formations typically fail. 

As shown in Figure \ref{fig:trajectories}, we further illustrate the trajectories of a five-robot team with different formations during navigation. We can see that our approach can control the robot team to navigate through narrow corridors while adaptively controlling the team formation in challenging scenarios, which indicates the effectiveness of our multi-robot navigation method with formation control and adaptation capabilities.

The quantitative results are shown in Table \ref {tab:formation-integrity}. 
DGNN, which focuses solely on goal achievement, performs poorly in CFI metrics due to its neglect of formation control. Although it achieves a 100\% success rate in circle and line formations, it only reaches 60\% in wedge formation and exhibits significant oscillation and deviation after goal completion. L$\&$F performs better than DGNN in CFI metrics, maintaining a rigid formation until narrow passages are encountered. However, its rigid formation fails to adapt to complex scenarios, leading to a 60\% success rate in wedge and line formations.
The baseline method outperforms previous approaches by achieving a 100\% success rate through effective obstacle avoidance. Nevertheless, its performance in CFI metrics is lower than AFOR's due to the absence of a spring-damper model, especially under high uncertainty. AFOR, with its focus on both formation control and adaptation, consistently delivers superior performance across all formations and uncertainty levels. It still maintains desired formation even as uncertainty decreases, showcasing robust adaptive control during navigation.


\begin{figure*}[htb]
\centering
\subfigure{\includegraphics[width=0.950\textwidth]{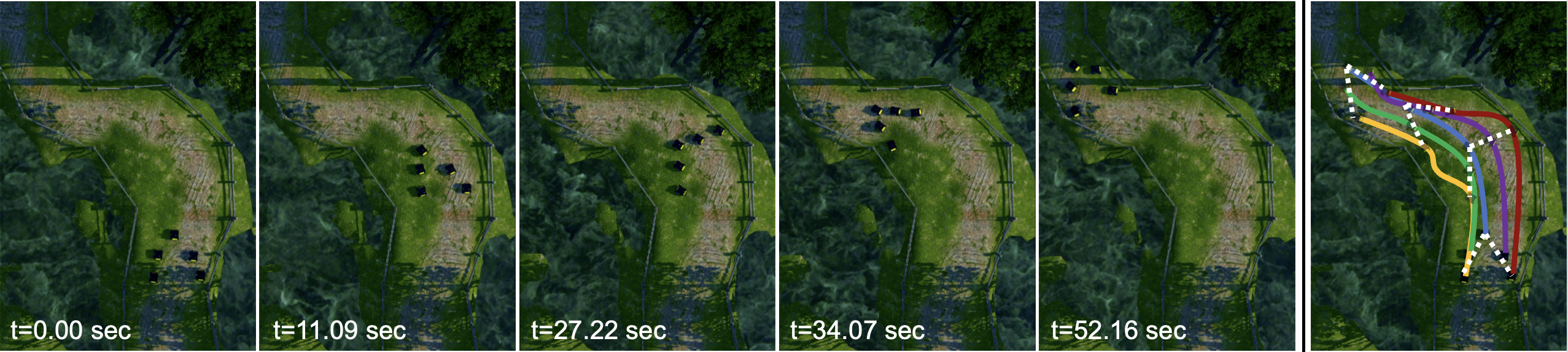}}
\vspace{-6pt}
\caption{Qualitative results on formation adaptation using a team of Husky robots in a high-fidelity multi-robot simulation that integrates Unity3D with ROS. The robots adaptively maintain a wedge formation when navigating through a long and narrow corridor. [More result videos are available on the project website.]}
\label{fig:3d}
\vspace{-12pt}
\end{figure*}

\begin{figure*}[h]
\centering
\subfigure[3 robots]{\includegraphics[height=4.5cm]{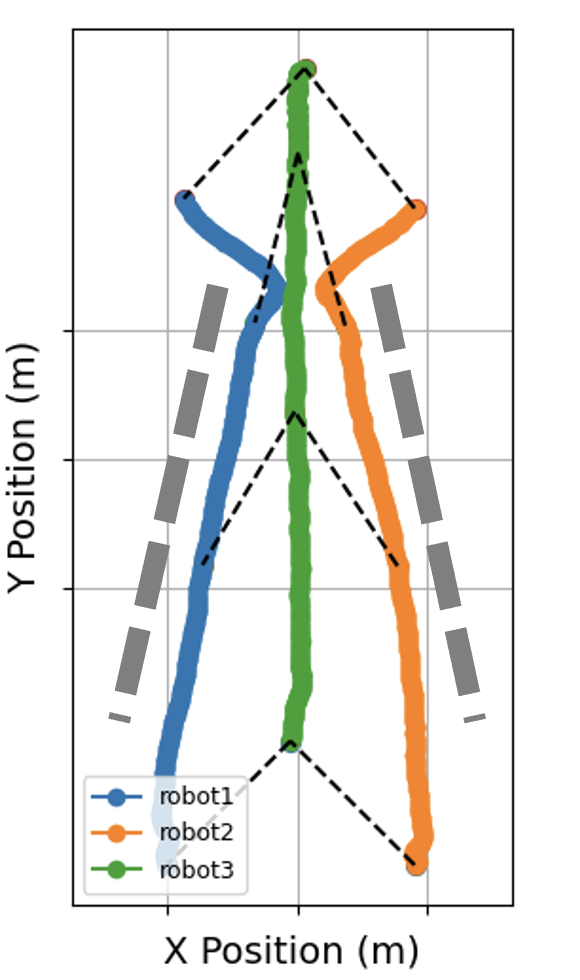}\label{fig:dis3}}
\hspace{5pt}
\subfigure[5 robots]{\includegraphics[height=4.5cm]{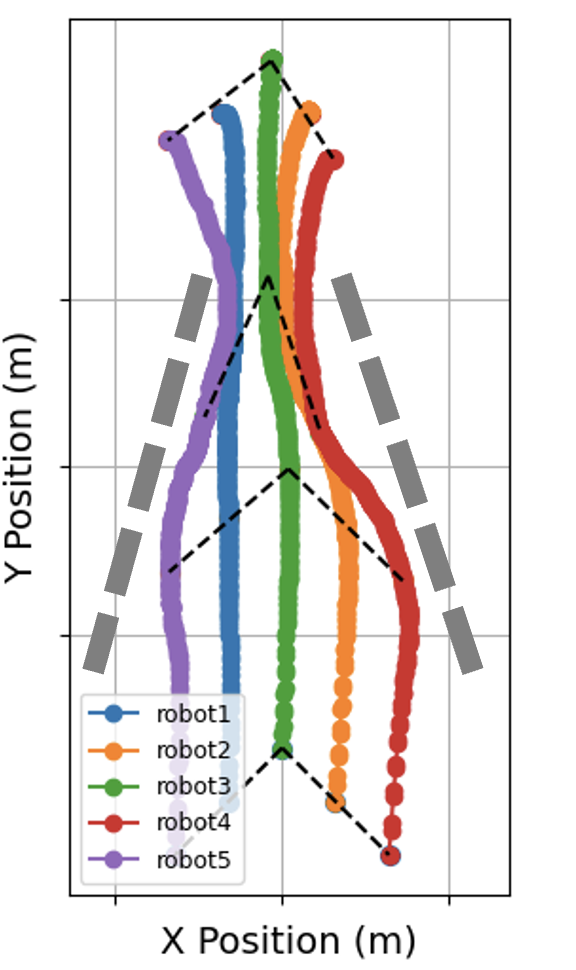}\label{fig:dis5}}
\hspace{5pt}
\subfigure[7 robots]{\includegraphics[height=4.5cm]{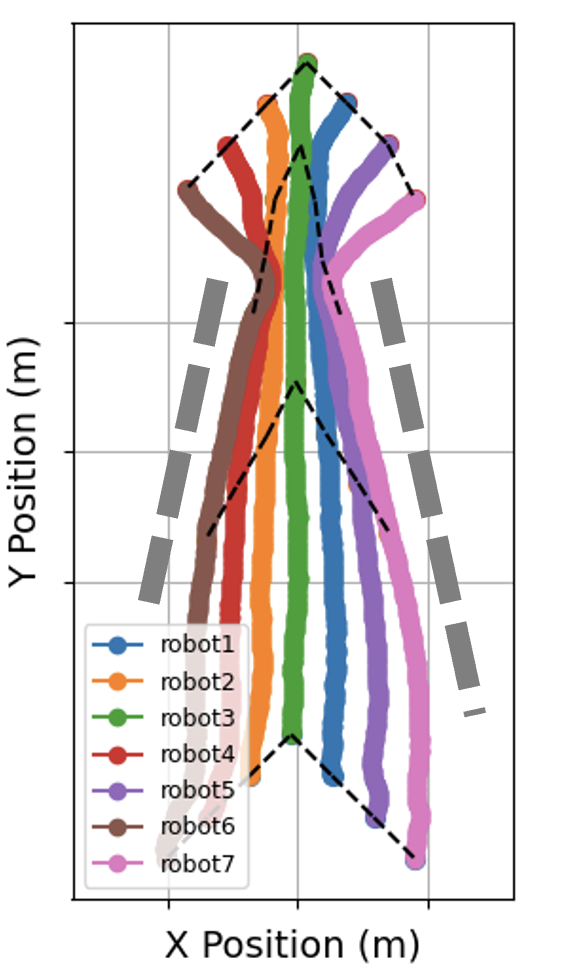}\label{fig:dis7}}
\hspace{5pt}
\subfigure[9 robots]{\includegraphics[height=4.5cm]{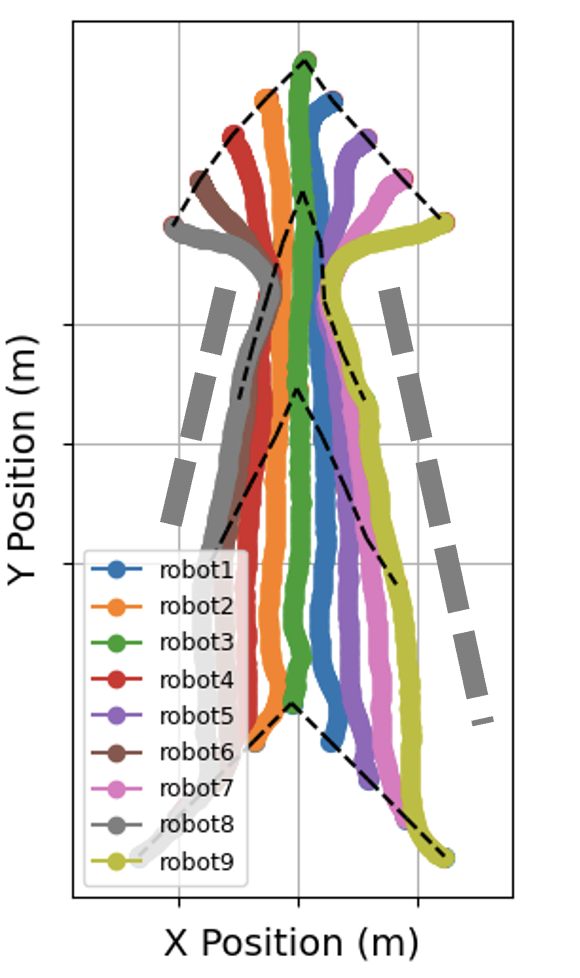}\label{fig:dis9}}
\hspace{5pt}
\subfigure[CFI values across different team sizes]{\includegraphics[height=4.5cm]{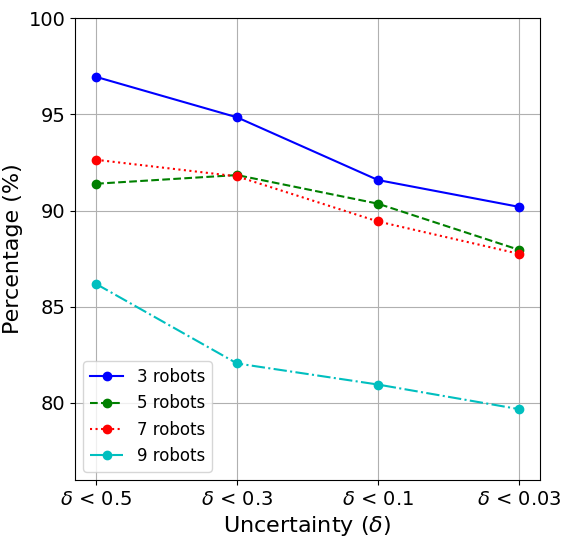}\label{fig:discompare}}
\vspace{-4pt}
\caption{Quantitative results on the AFOR's generalizability to varied team sizes. Figures (a)-(d) shows movement trajectories of a team of 3-9 robots navigating through a narrow corridor while maintaining a wedge formation. Figure (e) illustrates changes in the CFI values given different team sizes and $\delta$ values.}
\label{fig:dis}
\vspace{-6pt}

\end{figure*}

\subsection{Case Studies on Real Robots and High-fidelity Simulations}
We validate AFOR using a physical multi-robot system that consists of five holonomic wheeled robots operating in three different formations. The team is controlled in a centralized way by a master computer equipped with an Intel i7 8-core CPU, 24GB of memory, and an NVIDIA TITAN RTX GPU. 
As shown in Figure \ref{fig:real}, all five robots maintain the original formation at the beginning. When the robot team reaches the narrow corridors, our approach adapts this formation to a more compact one, maximizing both team integrity and the efficient use of available space. After navigating through the narrow passage, the robots are able to reach their goal successfully. The real-world experiments demonstrate the effectiveness of our approach in enabling formation control and adaptation on real robot platforms in real-world complex environments.

To further evaluate AFOR in a realistic outdoor environment, we conduct experiments using a high-fidelity simulator that incorporates the Unity3D engine with ROS for multi-robot control. We deploy five differential-drive Husky robots in wedge formation. The actions $\mathbf{a}_i$ with linear velocities are directly mapped to the differential-drive robot's wheel velocities to follow the same trajectory. 
As shown in Figure \ref{fig:3d}, we have an extremely complex scenario of multi-robot navigation. This includes navigating along curved trajectories and adapting team formations to pass through multiple narrow areas along the path. Given the qualitative results, we can see that our approach enables the robot team to smoothly and successfully navigate through the curved trajectory. In addition, our approach can continuously adjust their formation throughout the process, as demonstrated at $t=11.09$ sec and $t=34.07$ sec.
\color{black}

\subsection{Discussion on AFOR's Generalizability}\label{sec:discussion}
We further study the generalizability of AFOR by evaluating formation adaptation with varying numbers of robots.
Figure \ref{fig:dis3}-\ref{fig:dis9} illustrate that our approach performs effectively with teams of $3$, $5$, $7$, and $9$ robots in a wedge formation. The trajectories in all scenarios demonstrate smooth formation adaptation capability during navigation, highlighting the effectiveness of our approach in formation control and adaptation across different team sizes.

Figure \ref{fig:discompare} presents the quantitative analysis of our approach with different team sizes based on the CFI metric. The results show that our approach maintains strong formation integrity when $\delta < 0.5$. As the number of robots increases, formation integrity slightly decreases but stabilizes when $\delta > 0.1$, indicating the robustness of our approach for multi-robot formation adaptation. For the large team size with $9$ robots,  our approach achieves at least 80$\%$ formation integrity. For small teams with $3$ robots, our approach can maintain above 90$\%$ integrity, indicating the effectiveness of our approach in maintaining team formation during navigation.

\section{Conclusion}\label{sec:conclusion}


In this paper, we propose AFOR to enable formation adaptation for coordinated multi-robot navigation in challenging scenarios that include narrow corridors.
AFOR is developed within a hierarchical learning framework, integrating a spring-damper model, to enable both team coordination and individual robot control. 
The upper level uses a GNN to perform adaptive coordination and information sharing,
while the lower level uses RL to learn individual robot control. 
Results from comprehensive experiments show that AFOR enables a team of robots to navigate with formation adaptation, and it outperforms the previous methods in coordinated multi-robot navigation. 
 Future work will focus on exploring decentralized execution to enhance scalability and robustness, and transitioning to 3D environments to tackle challenges posed by uneven terrain in more realistic field settings.

\bibliographystyle{IEEEtran}
\bibliography{ref}

\begin{thebibliography}{10}
\providecommand{\url}[1]{#1}
\csname url@rmstyle\endcsname
\providecommand{\newblock}{\relax}
\providecommand{\bibinfo}[2]{#2}
\providecommand\BIBentrySTDinterwordspacing{\spaceskip=0pt\relax}
\providecommand\BIBentryALTinterwordstretchfactor{4}
\providecommand\BIBentryALTinterwordspacing{\spaceskip=\fontdimen2\font plus
\BIBentryALTinterwordstretchfactor\fontdimen3\font minus
  \fontdimen4\font\relax}
\providecommand\BIBforeignlanguage[2]{{%
\expandafter\ifx\csname l@#1\endcsname\relax
\typeout{** WARNING: IEEEtran.bst: No hyphenation pattern has been}%
\typeout{** loaded for the language `#1'. Using the pattern for}%
\typeout{** the default language instead.}%
\else
\language=\csname l@#1\endcsname
\fi
#2}}

\bibitem{drew2021multi}
D.~S. Drew, ``Multi-agent systems for search and rescue applications,''
  \emph{Current Robotics Reports}, vol.~2, pp. 189--200, 2021.

\bibitem{queralta2020collaborative}
J.~P. Queralta, J.~Taipalmaa, B.~C. Pullinen, V.~K. Sarker, T.~N. Gia,
  H.~Tenhunen, M.~Gabbouj, J.~Raitoharju, and T.~Westerlund, ``Collaborative
  multi-robot search and rescue: Planning, coordination, perception, and active
  vision,'' \emph{IEEE Access}, vol.~8, pp. 191\,617--191\,643, 2020.

\bibitem{hu2022multi}
J.~Hu, W.~Liu, H.~Zhang, J.~Yi, and Z.~Xiong, ``Multi-robot object transport
  motion planning with a deformable sheet,'' \emph{IEEE Robotics and Automation
  Letters}, vol.~7, no.~4, pp. 9350--9357, 2022.

\bibitem{gao2023collaborative}
P.~Gao, S.~Siva, A.~Micciche, and H.~Zhang, ``Collaborative scheduling with
  adaptation to failure for heterogeneous robot teams,'' in \emph{IEEE
  International Conference on Robotics and Automation}, 2023.

\bibitem{han2020cooperative}
R.~Han, S.~Chen, and Q.~Hao, ``Cooperative multi-robot navigation in dynamic
  environment with deep reinforcement learning,'' in \emph{IEEE International
  Conference on Robotics and Automation}, 2020.

\bibitem{indelman2018cooperative}
V.~Indelman, ``Cooperative multi-robot belief space planning for autonomous
  navigation in unknown environments,'' \emph{Autonomous Robots}, vol.~42, pp.
  353--373, 2018.

\bibitem{tang2020multi}
B.~Tang, K.~Xiang, M.~Pang, and Z.~Zhanxia, ``Multi-robot path planning using
  an improved self-adaptive particle swarm optimization,'' \emph{International
  Journal of Advanced Robotic Systems}, vol.~17, no.~5, p. 1729881420936154,
  2020.

\bibitem{galceran2013survey}
E.~Galceran and M.~Carreras, ``A survey on coverage path planning for
  robotics,'' \emph{Robotics and Autonomous Systems}, vol.~61, no.~12, pp.
  1258--1276, 2013.

\bibitem{thabit2018multi}
S.~Thabit and A.~Mohades, ``Multi-robot path planning based on multi-objective
  particle swarm optimization,'' \emph{IEEE Access}, vol.~7, pp. 2138--2147,
  2018.

\bibitem{reily2020leading}
B.~Reily, C.~Reardon, and H.~Zhang, ``Leading multi-agent teams to multiple
  goals while maintaining communication,'' in \emph{Robotics Science and
  Systems}, 2020.

\bibitem{chen2017socially}
Y.~F. Chen, M.~Everett, M.~Liu, and J.~P. How, ``Socially aware motion planning
  with deep reinforcement learning,'' in \emph{IEEE/RSJ International
  Conference on Intelligent Robots and Systems}, 2017.

\bibitem{yang2020multi}
Y.~Yang, L.~Juntao, and P.~Lingling, ``Multi-robot path planning based on a
  deep reinforcement learning dqn algorithm,'' \emph{CAAI Transactions on
  Intelligence Technology}, vol.~5, no.~3, pp. 177--183, 2020.

\bibitem{li2020graph}
Q.~Li, F.~Gama, A.~Ribeiro, and A.~Prorok, ``Graph neural networks for
  decentralized multi-robot path planning,'' in \emph{IEEE/RSJ International
  Conference on Intelligent Robots and Systems}, 2020.

\bibitem{zhang2023neural}
S.~Zhang, K.~Garg, and C.~Fan, ``Neural graph control barrier functions guided
  distributed collision-avoidance multi-agent control,'' in \emph{Conference on
  Robot Learning}, 2023.

\bibitem{he2022dynamic}
Z.~He, C.~Liu, X.~Chu, R.~R. Negenborn, and Q.~Wu, ``Dynamic anti-collision
  a-star algorithm for multi-ship encounter situations,'' \emph{Applied Ocean
  Research}, vol. 118, p. 102995, 2022.

\bibitem{sun2019novel}
G.~Sun, R.~Zhou, B.~Di, Z.~Dong, and Y.~Wang, ``A novel cooperative path
  planning for multi-robot persistent coverage with obstacles and coverage
  period constraints,'' \emph{Sensors}, vol.~19, no.~9, p. 1994, 2019.

\bibitem{bai2019distributed}
X.~Bai, W.~Yan, M.~Cao, and D.~Xue, ``Distributed multi-vehicle task assignment
  in a time-invariant drift field with obstacles,'' \emph{IET Control Theory \&
  Applications}, vol.~13, no.~17, pp. 2886--2893, 2019.

\bibitem{mac2017hierarchical}
T.~T. Mac, C.~Copot, D.~T. Tran, and R.~De~Keyser, ``A hierarchical global path
  planning approach for mobile robots based on multi-objective particle swarm
  optimization,'' \emph{Applied Soft Computing}, vol.~59, pp. 68--76, 2017.

\bibitem{peng2015multi}
J.-H. Peng, I.-H. Li, Y.-H. Chien, C.-C. Hsu, and W.-Y. Wang, ``Multi-robot
  path planning based on improved d* lite algorithm,'' in \emph{IEEE 12th
  International Conference on Networking, Sensing and Control}, 2015.

\bibitem{cappello2020hybrid}
D.~Cappello, S.~Garcin, Z.~Mao, M.~Sassano, A.~Paranjape, and T.~Mylvaganam,
  ``A hybrid controller for multi-agent collision avoidance via a differential
  game formulation,'' \emph{IEEE Transactions on Control Systems Technology},
  vol.~29, no.~4, pp. 1750--1757, 2020.

\bibitem{alonso2017multi}
J.~Alonso-Mora, S.~Baker, and D.~Rus, ``Multi-robot formation control and
  object transport in dynamic environments via constrained optimization,''
  \emph{The International Journal of Robotics Research}, vol.~36, no.~9, pp.
  1000--1021, 2017.

\bibitem{alonso2016distributed}
J.~Alonso-Mora, E.~Montijano, M.~Schwager, and D.~Rus, ``Distributed
  multi-robot formation control among obstacles: A geometric and optimization
  approach with consensus,'' in \emph{IEEE International Conference on Robotics
  and Automation}, 2016.

\bibitem{koung2021cooperative}
D.~Koung, O.~Kermorgant, I.~Fantoni, and L.~Belouaer, ``Cooperative multi-robot
  object transportation system based on hierarchical quadratic programming,''
  \emph{IEEE Robotics and Automation Letters}, vol.~6, no.~4, pp. 6466--6472,
  2021.

\bibitem{di2021multi}
G.~A. Di~Caro and A.~W.~Z. Yousaf, ``Multi-robot informative path planning
  using a leader-follower architecture,'' in \emph{IEEE International
  Conference on Robotics and Automation}, 2021.

\bibitem{yu2019formation}
H.~Yu, P.~Shi, C.-C. Lim, and D.~Wang, ``Formation control for multi-robot
  systems with collision avoidance,'' \emph{International Journal of Control},
  vol.~92, no.~10, pp. 2223--2234, 2019.

\bibitem{wu2022leader}
T.~Wu, K.~Xue, and P.~Wang, ``Leader-follower formation control of usvs using
  apf-based adaptive fuzzy logic nonsingular terminal sliding mode control
  method,'' \emph{Journal of Mechanical Science and Technology}, vol.~36,
  no.~4, pp. 2007--2018, 2022.

\bibitem{xiao2019leader}
H.~Xiao and C.~P. Chen, ``Leader-follower consensus multi-robot formation
  control using neurodynamic-optimization-based nonlinear model predictive
  control,'' \emph{IEEE Access}, vol.~7, pp. 43\,581--43\,590, 2019.

\bibitem{ying2015leader}
Z.~Ying and L.~Xu, ``Leader-follower formation control and obstacle avoidance
  of multi-robot based on artificial potential field,'' in \emph{The 27th
  Chinese Control and Decision Conference}, 2015.

\bibitem{bai2021learning}
C.~Bai, P.~Yan, W.~Pan, and J.~Guo, ``Learning-based multi-robot formation
  control with obstacle avoidance,'' \emph{IEEE Transactions on Intelligent
  Transportation Systems}, vol.~23, no.~8, pp. 11\,811--11\,822, 2021.

\bibitem{benzerrouk2014stable}
A.~Benzerrouk, L.~Adouane, and P.~Martinet, ``Stable navigation in formation
  for a multi-robot system based on a constrained virtual structure,''
  \emph{Robotics and Autonomous Systems}, vol.~62, no.~12, pp. 1806--1815,
  2014.

\bibitem{alonso2019distributed}
J.~Alonso-Mora, E.~Montijano, T.~N{\"a}geli, O.~Hilliges, M.~Schwager, and
  D.~Rus, ``Distributed multi-robot formation control in dynamic
  environments,'' \emph{Autonomous Robots}, vol.~43, pp. 1079--1100, 2019.

\bibitem{roy2019virtual}
D.~Roy, A.~Chowdhury, M.~Maitra, and S.~Bhattacharya, ``Virtual region based
  multi-robot path planning in an unknown occluded environment,'' in
  \emph{IEEE/RSJ International Conference on Intelligent Robots and Systems},
  2019.

\bibitem{abujabal2023comprehensive}
N.~Abujabal, R.~Fareh, S.~Sinan, M.~Baziyad, and M.~Bettayeb, ``A comprehensive
  review of the latest path planning developments for multi-robot formation
  systems,'' \emph{Robotica}, vol.~41, no.~7, pp. 2079--2104, 2023.

\bibitem{roy2018multi}
D.~Roy, A.~Chowdhury, M.~Maitra, and S.~Bhattacharya, ``Multi-robot virtual
  structure switching and formation changing strategy in an unknown occluded
  environment,'' in \emph{IEEE/RSJ International Conference on Intelligent
  Robots and Systems}, 2018.

\bibitem{han2022reinforcement}
R.~Han, S.~Chen, S.~Wang, Z.~Zhang, R.~Gao, Q.~Hao, and J.~Pan, ``Reinforcement
  learned distributed multi-robot navigation with reciprocal velocity obstacle
  shaped rewards,'' \emph{IEEE Robotics and Automation Letters}, vol.~7, no.~3,
  pp. 5896--5903, 2022.

\bibitem{hacene2021behavior}
N.~Hacene and B.~Mendil, ``Behavior-based autonomous navigation and formation
  control of mobile robots in unknown cluttered dynamic environments with
  dynamic target tracking,'' \emph{International Journal of Automation and
  Computing}, vol.~18, no.~5, pp. 766--786, 2021.

\bibitem{bischoff2013hierarchical}
B.~Bischoff, D.~Nguyen-Tuong, I.~Lee, F.~Streichert, A.~Knoll, \emph{et~al.},
  ``Hierarchical reinforcement learning for robot navigation,'' in
  \emph{Proceedings of The European Symposium on Artificial Neural Networks,
  Computational Intelligence And Machine Learning}, 2013.

\bibitem{jin2021hierarchical}
Y.~Jin, S.~Wei, J.~Yuan, and X.~Zhang, ``Hierarchical and stable multiagent
  reinforcement learning for cooperative navigation control,'' \emph{IEEE
  Transactions on Neural Networks and Learning Systems}, vol.~34, no.~1, pp.
  90--103, 2021.

\bibitem{yang2018hierarchical}
Z.~Yang, K.~Merrick, L.~Jin, and H.~A. Abbass, ``Hierarchical deep
  reinforcement learning for continuous action control,'' \emph{IEEE
  Transactions on Neural Networks and Learning Systems}, vol.~29, no.~11, pp.
  5174--5184, 2018.

\bibitem{chang2023hierarchical}
L.~Chang, L.~Shan, W.~Zhang, and Y.~Dai, ``Hierarchical multi-robot navigation
  and formation in unknown environments via deep reinforcement learning and
  distributed optimization,'' \emph{Robotics and Computer-Integrated
  Manufacturing}, vol.~83, p. 102570, 2023.

\bibitem{wohlke2021hierarchies}
J.~W{\"o}hlke, F.~Schmitt, and H.~van Hoof, ``Hierarchies of planning and
  reinforcement learning for robot navigation,'' in \emph{IEEE International
  Conference on Robotics and Automation}, 2021.

\bibitem{hu2020voronoi}
J.~Hu, H.~Niu, J.~Carrasco, B.~Lennox, and F.~Arvin, ``Voronoi-based
  multi-robot autonomous exploration in unknown environments via deep
  reinforcement learning,'' \emph{IEEE Transactions on Vehicular Technology},
  vol.~69, no.~12, pp. 14\,413--14\,423, 2020.

\bibitem{zhu2022hierarchical}
W.~Zhu and M.~Hayashibe, ``A hierarchical deep reinforcement learning framework
  with high efficiency and generalization for fast and safe navigation,''
  \emph{IEEE Transactions on Industrial Electronics}, vol.~70, no.~5, pp.
  4962--4971, 2022.

\bibitem{pal2021learning}
A.~Pal, Y.~Qiu, and H.~Christensen, ``Learning hierarchical relationships for
  object-goal navigation,'' in \emph{Conference on Robot Learning}, 2021.

\bibitem{liu2020robot}
L.~Liu, D.~Dugas, G.~Cesari, R.~Siegwart, and R.~Dub{\'e}, ``Robot navigation
  in crowded environments using deep reinforcement learning,'' in
  \emph{IEEE/RSJ International Conference on Intelligent Robots and Systems},
  2020.

\bibitem{blumenkamp2022framework}
J.~Blumenkamp, S.~Morad, J.~Gielis, Q.~Li, and A.~Prorok, ``A framework for
  real-world multi-robot systems running decentralized {GNN}-based policies,''
  in \emph{IEEE International Conference on Robotics and Automation}, 2022.

\bibitem{rychlewski1984hooke}
J.~Rychlewski, ``On hooke's law,'' \emph{Journal of Applied Mathematics and
  Mechanics}, vol.~48, no.~3, pp. 303--314, 1984.

\bibitem{Godard_2019_ICCV}
C.~Godard, O.~Mac~Aodha, M.~Firman, and G.~J. Brostow, ``Digging into
  self-supervised monocular depth estimation,'' in \emph{Proceedings of the
  IEEE/CVF International Conference on Computer Vision}, 2019.

\end{thebibliography}
\end{document}